\documentclass[10pt,twocolumn,letterpaper]{article}

\usepackage{wacv}
\usepackage{times}
\usepackage{epsfig}
\usepackage{graphicx}
\usepackage{amsmath}
\usepackage{amssymb}
\usepackage{multirow}
\usepackage{bbm}

\def\wacvPaperID{463} 

\ifwacvfinal
\def\assignedStartPage{1} 
\fi

\ifwacvfinal
\usepackage[breaklinks=true,bookmarks=false]{hyperref}
\else
\usepackage[pagebackref=true,breaklinks=true,colorlinks,bookmarks=false]{hyperref}
\fi

\ifwacvfinal
\else
\fi

\begin{document}

\title{Objectness-Aware Few-Shot Semantic Segmentation}

\author{Yinan Zhao$^*$,
Brian Price$^+$,
Scott Cohen$^+$,
Danna Gurari$^*$\\
{\small $~^*$ University of Texas at Austin,} 
{\small $~^+$ Adobe Research}\\
{\tt\small yinanzhao@utexas.edu, \{bprice,scohen\}@adobe.com, danna.gurari@ischool.utexas.edu} 
}

\maketitle

\begin{abstract}
Few-shot semantic segmentation models aim to segment images after learning from only a few annotated examples. A key challenge for them is how to avoid overfitting because limited training data is available. While prior works usually limited the overall model capacity to alleviate overfitting, this hampers segmentation accuracy. We demonstrate how to increase overall model capacity to achieve improved performance, by introducing objectness, which is class-agnostic and so not prone to overfitting, for complementary use with class-specific features. Extensive experiments demonstrate the versatility of our simple approach of introducing objectness for different base architectures that rely on different data loaders and training schedules (DENet~\cite{liu2020dynamic}, PFENet~\cite{tian2020prior}) as well as with different backbone models (ResNet-50, ResNet-101 and HRNetV2-W48). Given only one annotated example of an unseen category, experiments show that our method outperforms state-of-art methods with respect to mIoU by at least 4.7\% and 1.5\% on PASCAL-$5^i$ and COCO-$20^i$ respectively. 
\end{abstract}
\section{Introduction}

\begin{figure*}[!th]
\centering
\includegraphics[width=0.95\textwidth]{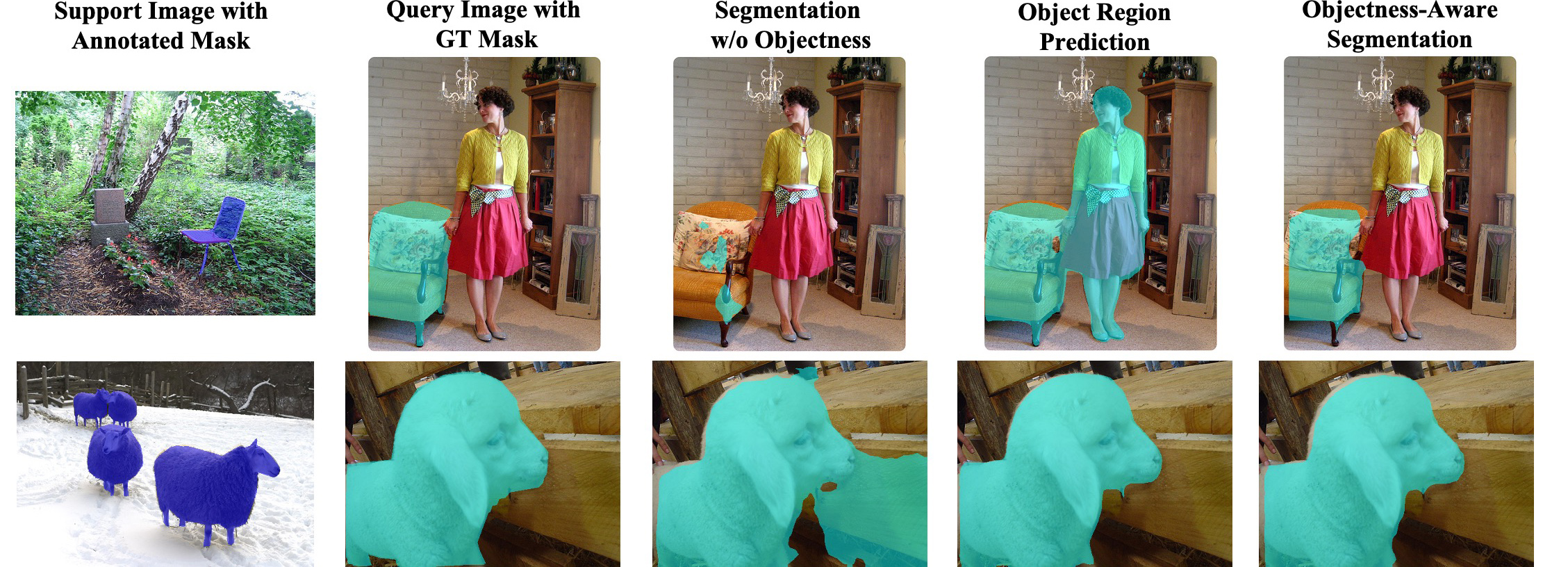}
\vspace{0pt}
\caption{One-shot semantic segmentation results for two images (one per row). Shown from left to right are: support image with segmentation mask, query image with ground truth segmentation mask, predicted segmentation of our implemented baseline without objectness, object regions predicted by our objectness module, and predicted segmentation of our method with objectness.  (Best viewed in color to observe the segmentation masks that are overlaid on all the images.)}
\label{fig:intro}
\end{figure*}

Semantic segmentation methods predict a category label for every pixel in an image. While there has been significant progress for this task using deep neural networks~\cite{long2015fully,chen2018encoder,zhao2017pyramid,chen2017rethinking,yuan2018ocnet,liu2015parsenet,yang2018denseaspp,zhao2018psanet,fu2019dual,fourure2017gridnet,sun2019high}, most methods require a large number of images with pixel-wise annotations for training. Yet, annotation collection is expensive and resulting models are constrained to predict only a pre-defined set of categories. Few-shot semantic segmentation methods, in contrast, learn to predict the segmentation mask of a novel object category in a \emph{query image} by observing $K$ images paired with segmentation masks of that category at test time. This task is called $K$-shot semantic segmentation and each image-mask pair is called a \emph{support image}.  

A promising trend for few-shot semantic segmentation methods~\cite{rakelly2018conditional,hu2019attention,zhang2018sg,zhang2019canet,zhang2019pyramid,nguyen2019feature,wang2020few,gairola2020simpropnet,liu2020crnet,tian2020prior} is to measure the feature similarity between every spatial location of the query image and the support images in order to determine which of its pixels belong to the same object category. Yet, such methods still struggle to perform well in all scenarios.  One scenario is when the target object category has a large appearance mismatch between the support image and query image.  For example, in row 1 in Figure~\ref{fig:intro}, the annotated chair in the support image (column 1) only matches part of the chair in the query image through feature comparison (column 3). Another scenario is when the query image has a cluttered background.  For example, in row 2 in Figure~\ref{fig:intro}, the annotated sheep (column 1) match not only the sheep in the query image, but also a background region with a similar texture to the target sheep category (column 3). 

The key challenge for few-shot semantic segmentation methods to generalize well across all scenarios is overfitting.  That is because only a few training examples of the target class are available. A common approach adopted by prior work~\cite{zhang2019canet,zhang2019pyramid,yang2020new} to alleviate overfitting is to \emph{limit} the capacity of models.  However, doing so hampers the segmentation accuracy. 

Our key contribution is an approach for increasing a model's performance by \emph{increasing} the model capacity, which was found non-trivial in prior few-shot segmentation work~\cite{zhang2019canet,zhang2019pyramid,yang2020new,tian2020prior}.  Our work is inspired by the observation that foreground objects, regardless of semantic categories, share some features that differentiate themselves from the background. Our method incorporates this prior knowledge by leveraging  \textbf{objectness}.  Doing so enables a separation of (1) \textbf{class-agnostic} features which capture the general object prior regardless of the object classes, and (2) \textbf{class-specific} features which encode the unique characteristics of the target object class. As we will show in Section~\ref{sec:experiments}, we find that class-specific features are prone to overfitting, while class-agnostic features are not. Such separation enables us to increase the overall model capacity without severe overfitting, by allocating more capacity to encode class-agnostic features while limiting the capacity for class-specific features.

Our proposed method consists of two key steps.  First, we train an objectness module to differentiate objects from background. Then we introduce an objectness-aware dense comparison module to match class-specific features between the query image and support images, while being aware of the class-agnostic objectness map predicted by our pretrained objectness module. 

Benefits of our method are exemplified in Figure~\ref{fig:intro}.  For the chair example (row 1), our objectness module predicts both the chair and person regions to be objects in the query image (column 4). The objectness-aware comparison module removes the person and keeps most of the chair region after observing class-specific features of the annotated chair in the support image (column 5). For the sheep example (row 2), our objectness module predicts an accurate object region for the sheep in the query image (column 4). The objectness-aware comparison module transfers the class-agnostic object prior to the final prediction so that the misclassified background regions (column 3) are correctly regarded as background in the predicted segmentation mask (column 5).  These examples illustrate our method's ability to handle appearance mismatches between the support and query image (row 1) and cluttered backgrounds (row 2).  Quantitative experiments (Section~\ref{sec:experiments}) further underscore the benefit of leveraging objectness, with our method yielding state-of-the-art results on two datasets (PASCAL-$5^i$ and COCO-$20^i$). They also demonstrate the versatility of our method when employed with different backbones for the objectness and feature extraction module, and with multiple base architectures (DENet~\cite{liu2020dynamic} and PFENet~\cite{tian2020prior}).
\section{Related Work}

\paragraph{Semantic Segmentation.} 
Deep convolutional neural networks have been the dominant solution for semantic segmentation since 2015 when a fully connected network (FCN)~\cite{long2015fully} was shown to yield significant improvements over hand-crafted features~\cite{he2004multiscale,shotton2009textonboost,ladicky2009associative}. Yet FCN~\cite{long2015fully}, and subsequent improvements to this framework~\cite{chen2018encoder,zhao2017pyramid,chen2017rethinking,yuan2018ocnet,liu2015parsenet,yang2018denseaspp,zhao2018psanet,fu2019dual,fourure2017gridnet,sun2019high} share a need for a large number of densely annotated images during training.  Unlike these methods, we focus on the few-shot scenario where there are limited number of annotated images for an object category.

\vspace{-1em}\paragraph{Few-Shot Learning.} 
Generally, this problem entails learning from a small amount of annotated data knowledge that can transfer to new classes. Often it has been studied for the classification task, with approaches incuding parameter prediction~\cite{bertinetto2016learning,wang2016learning}, optimization model learning~\cite{finn2017model,ravi2016optimization} and metric learning~\cite{koch2015siamese,snell2017prototypical,sung2018learning,vinyals2016matching}. We similarly adopt a metric learning based approach by using feature similarity comparison, but we instead focus on it in a dense form for tackling semantic segmentation.

\vspace{-1em}\paragraph{Few-Shot Semantic Segmentation.}
Since 2017, when few-shot learning was introduced for semantic segmentation with a metric learning approach~\cite{shaban2017one}, subsequent approaches have included using meta-learning~\cite{tian2019differentiable}, adopting prototypical networks~\cite{dong2018few,wang2019panet,yang2020prototype,liu2020part,snell2017prototypical}, and imprinting weights~\cite{siam2019amp}. Given the widespread success of methods that adopt a metric learning approach ~\cite{rakelly2018conditional,hu2019attention,zhang2018sg,zhang2019canet,zhang2019pyramid,nguyen2019feature,wang2020few,gairola2020simpropnet,liu2020crnet,tian2020prior}, we similarly build upon this framework by designing a method that compares features describing the query image and support images. Unlike prior work, we demonstrate how to \emph{increase} model capacity by introducing objectness.  Our experiments demonstrate its benefit both for 1-shot and 5-shot semantic segmentation.

\vspace{-1em}\paragraph{Foreground Object Segmentation.}
The task of predicting a binary mask of object regions in an image in a category-independent manner is a fundamental problem facilitating a wide range of vision applications such as scene understanding~\cite{bharath2013scalable,ramesh2019scalable} and caption generation~\cite{cornia2018paying,zhou2019re}. While earlier approaches rely on low-level hand-crafted features~\cite{russell2006using,endres2010category,zitnick2014edge,arbelaez2014multiscale,carreira2011cpmc,krahenbuhl2014geodesic}, the status quo has become deep convolutional neural network (CNN) based approaches~\cite{kuo2015deepbox,xiong2018pixel,zeng2019wsod2,zhang2018mixed}.  To our knowledge, our work is the first to demonstrate the benefit of foreground object segmentation for tackling few-shot semantic segmentation.  


\begin{figure*}[!t]
\centering
\includegraphics[width=1.0\textwidth]{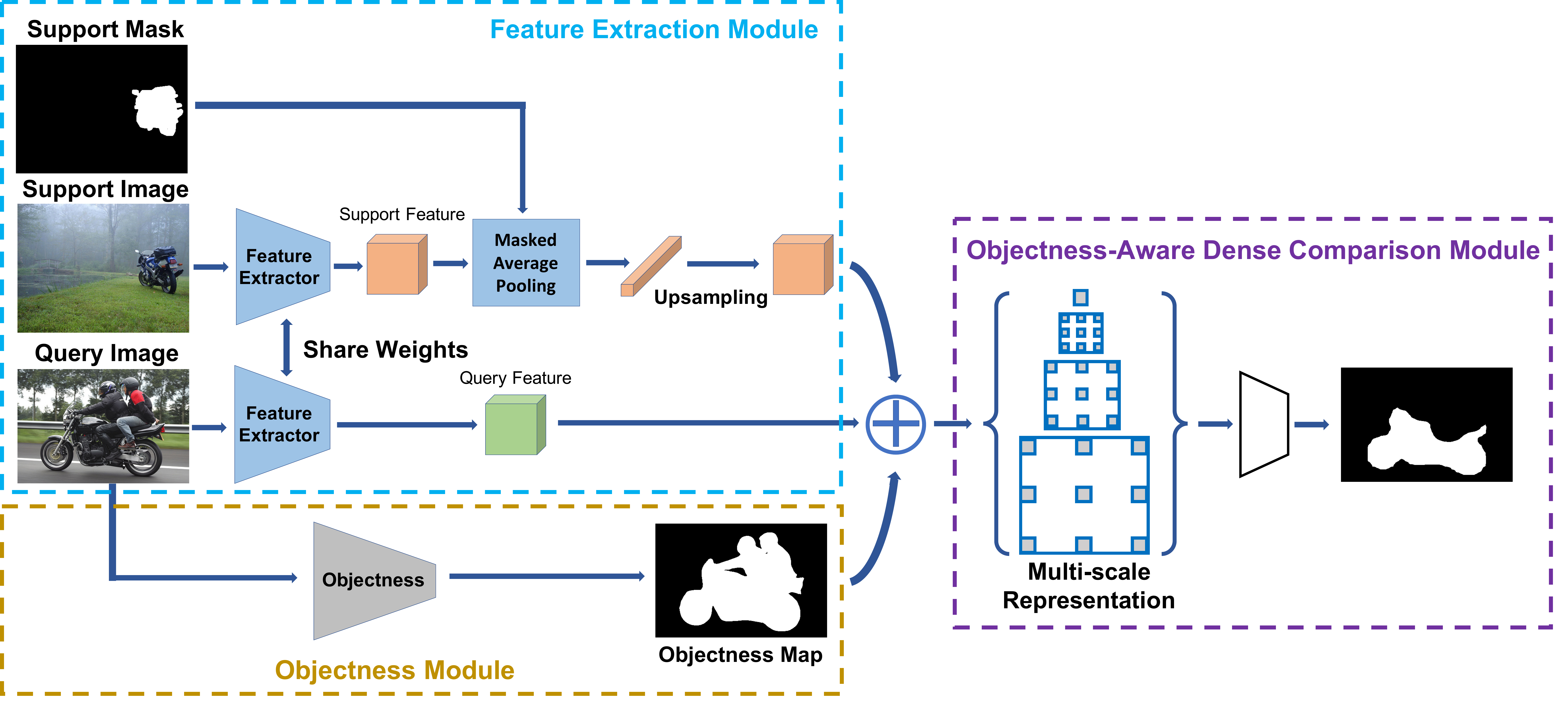}
\vspace{-12pt}
\caption{Our proposed framework for objectness-aware few-shot semantic segmentation. It consists of three modules: feature extraction module, objectness module and objectness-aware dense comparison module.}
\label{fig:method}
\end{figure*}

\section{Method}
We now introduce our few-shot semantic segmentation method that can \emph{increase} capacity \emph{without severe overfitting}.  The key elements of our method are the embedding of an objectness module in the model architecture (Section~\ref{sec:framework}) paired with independent training of this module from the class-specific components (Section~\ref{sec:implementation}).

\subsection{Problem Definition}
\label{sec:problem}
Let $\mathcal{D}_{\text{train}}$ and $\mathcal{D}_{\text{test}}$ represent two image sets containing non-overlapping object category sets $\mathcal{C}_{\text{train}}$ and $\mathcal{C}_{\text{test}}$ ($\mathcal{C}_{\text{train}}\cap\mathcal{C}_{\text{test}}=\emptyset$). Both the training set $\mathcal{D}_{\text{train}}$ and test set $\mathcal{D}_{\text{test}}$ consist of \textit{episodes}, with each consisting of a query image $\mathcal{Q}$ and a set $\mathcal{S}$ of support images representing the novel target category. Formally, $\mathcal{D}_{\text{train}}=\{(\mathcal{Q}_i^{train}, \mathcal{S}_i^{train})\}_{i=1}^{N_{train}}$ and $\mathcal{D}_{\text{test}}=\{(\mathcal{Q}_j^{test}, \mathcal{S}_j^{test})\}_{j=1}^{N_{test}}$. Each support image set $\mathcal{S}$ contains $K$ support images together with their annotated foreground masks of the object category. Formally, $\mathcal{S}=\{I_k, M_k\}_{k=1}^{K}$, where $I_k \in \mathbb{R}^{H_k \times W_k \times 3}$ is an RGB image and $M_k \in \mathbb{R}^{H_k \times W_k}$ is a binary mask denoting the annotation for the object category. 

We follow the same training and testing protocols employed by prior work~\cite{shaban2017one,rakelly2018conditional,dong2018few,siam2019amp,zhang2018sg,wang2019panet,zhang2019canet,zhang2019pyramid,nguyen2019feature,yang2020new}.  We randomly sample episodes from $\mathcal{D}_{\text{train}}$ to train a model and evaluate the trained model on $\mathcal{D}_{\text{test}}$ across all the testing episodes. The main challenge is how to make the knowledge learned from $\mathcal{D}_{\text{train}}$ generalize to $\mathcal{D}_{\text{test}}$, given that $\mathcal{D}_{\text{test}}$ represents a disjoint category set  $\mathcal{C}_{\text{test}}$.  

\subsection{Proposed Framework}
\label{sec:framework}
Our framework, summarized in Figure~\ref{fig:method}, consists of three modules: an objectness module,  feature extraction module, and objectness-aware dense comparison module. The objectness module produces a class-agnostic objectness map for the query image. In parallel, the feature extractor extracts representations for the query image and support images. Finally, the objectness-aware comparison module augments consideration of the objectness map to spatially compare class-specific features of the query and support image. We elaborate about each module below.

\subsubsection{Objectness Module}
\label{sec:saliency}
To capture the class-agnostic object prior, we design an objectness module $\mathcal{F}$ to produce an objectness map for the query image.  The ground truth object mask for a query image $I_Q$, $O_{I_Q}\in\{0,1\}^{H \times W}$, is a binary mask with 1 denoting object and 0 denoting background.  Our method takes as input a query image $I_Q$ and produces a map $\hat O_{I_Q}=\mathcal{F}(I_Q) \in \mathbb{R}^{H \times W}$, where each value denotes the probability that each spatial location is part of an object. 

Towards increasing overall model capacity, we use the parameter-heavy, strong segmentation model HRNetV2-W48~\cite{sun2019high} as the objectness module in our baseline model. To demonstrate the general-purpose benefit of objectness, we will also show our method's performance when using additional less parameter-heavy backbones (i.e., ResNet-50 and ResNet-101~\cite{he2016deep}) in Section~\ref{sec:ablation} as well as when augmenting our objectness module to existing state-of-art few-shot segmentation architectures (i.e., PFENet~\cite{tian2020prior} and DENet~\cite{liu2020dynamic}) in Sections~\ref{sec:experiment_pascal} and \ref{sec:experiment_coco}.

\subsubsection{Feature Extraction Module}
\label{sec:extraction}
The feature extraction module extracts representations of both the query image and support images.  The goal is for it to generate a compact class-specific feature for the target object category in the support images that can be used to find matching objects in the query image.  

Initially, the same feature extractor is applied to the query image and support images to generate a 1,536-channel feature map for each image. For the feature extractor, like prior work~\cite{wang2019panet,zhang2019canet,yang2020new}, we choose a backbone model that is pretrained on ImageNet~\cite{deng2009imagenet} to extract semantic image features. To support fair comparison to prior works~\cite{tian2020prior, yang2020new,nguyen2019feature,zhang2019canet,zhang2019pyramid}, which commonly use ResNet-50 as the feature extraction backbone, we also use in our baseline model ResNet-50 (Section~\ref{sec:experiment_pascal} and \ref{sec:experiment_coco}). To evaluate the influence of different backbones, we will also show our method's performance with other backbones (i.e., ResNet-101~\cite{he2016deep} and HRNetV2-W48~\cite{sun2019high}) in Section~\ref{sec:ablation}.  Then, we include a linear transformation to project the feature to a 256-dimensional feature space in order to reduce redundant dimensions. The linear transformation is implemented as a $1\times1$ convolutional layer. 

Next, for the support image only, we employ the \textit{masked average pooling} operation~\cite{zhang2018sg,zhang2019canet,wang2019panet} to extract a global vector for the target category.  By doing so, the model acquires a feature that represents the target object category while excluding other categories and background in the support image, because it only pools features in the region of the target object category.  Mathematically, masked average pooling is achieved as follows:
$$v_{\mathcal{S}}=\frac{1}{K}\sum\limits_{i=1}^K \frac{\sum\limits_{(h,w)}F_i^{(h,w)}\mathbbm{1}[{M_i^{(h,w)}}=1]}{\sum\limits_{(h,w)}\mathbbm{1}[{M_i^{(h,w)}}=1]}$$
where $(h,w)$ indexes the spatial locations, $K$ is the number of support images, $F_i$ and $M_i$ denote the feature map and ground truth mask of $i_{th}$ support image respectively and $\mathbbm{1}[\cdot]$ is an indicator function.

Finally, features extracted from the query and support images are concatenated.  To enable this concatenation, we upsample the global vector $v_{\mathcal{S}}$ describing the support image to the same spatial resolution as the query feature. Consequently, we have a repeated support feature at each location.

\subsubsection{Objectness-Aware Dense Comparison}
\label{sec:dense_comp}
The objectness-aware dense comparison module takes as input a concatenation of the class-specific query image features extracted by the feature extractor (described in Section~\ref{sec:extraction}), class-specific feature of the target category in the support image (described in Section~\ref{sec:extraction}), and class-agnostic objectness map produced by our objectness module (described in Section~\ref{sec:saliency}) in order to predict the segmentation mask in the query image. The concatenated input is passed through a multi-scale dense comparison module to produce a multi-scale representation.
Our baseline model embeds \textit{Feature Enrichment Module (FEM)}~\cite{tian2020prior}, since it is reported to achieve state-of-art performance.  Towards demonstrating the versatility of objectness in different achitectures, we also consider a variant that uses \textit{Atrous Spatial Pyramid Pooling Module (ASPP)} ~\cite{chen2017rethinking}.  Both \textit{ASPP} and \textit{FEM} consist of four parallel branches, where each produces a different representation for each scale. \textit{FEM} provides inter-scale interaction in the module as well. The output feature maps from the four scales are concatenated together and processed by two additional convolution layers before predicting the segmentation of the target category.

\subsection{Training}
\label{sec:implementation}
We train the different modules independently, as described below.  It will be demonstrated in Section~\ref{sec:experiments} that training the modules independently rather than jointly is a critical ingredient behind successfully increasing our model's capacity without severe overfitting.

First, we train the objectness module to encode class-agnostic features of the image. We initialize the objectness backbone with ImageNet~\cite{deng2009imagenet} pretrained weights and train it with the standard cross-entropy loss $Loss(O_{I_Q},\mathcal{F}(I_Q))$ between the ground truth binary mask $O_{I_Q}$ and our predicted objectness map $\mathcal{F}(I_Q)$. We derive the training set for objectness $\mathcal{D}_{\text{train}}^{\text{object}}$ from the training set for semantic segmentation $\mathcal{D}_{\text{train}}$, where $\mathcal{D}_{\text{train}}$ consists of images with semantic segmentation annotations. We treat all semantic categories except background as the foreground object class. Note that $\mathcal{D}_{\text{train}}^{\text{object}}$ contains no objects from $\mathcal{C}_{\text{test}}$. Therefore, the objectness module never observes any testing category in training. 

Next, the objectness-aware dense comparison module is trained with the objectness module and the pretrained feature extractor fixed.  The feature extractor module is fixed with the ImageNet~\cite{deng2009imagenet} pretrained weights to limit the number of learnable parameters for class-specific features.  The comparison module is trained using standard cross-entropy loss between the predicted segmentation mask and ground truth segmentation for the target category.

\section{Experiments}
\label{sec:experiments}
We now evaluate our proposed method in few-shot semantic segmentation and compare it to related baselines. We conduct experiments with two few-shot segmentation datasets, which are discussed in Sections \ref{sec:experiment_pascal} and \ref{sec:experiment_coco} respectively. We also conduct studies to motivate the design of our method and demonstrate its versatility when employed with different feature extraction backbones (Section~\ref{sec:ablation}).

\begin{table*}[!t]
\begin{center}
\begin{tabular}{| c  l | c  c  c  c | c | c | c  c  c  c | c | c |}
\hline
&\multirow{3}{*}{Method}&\multicolumn{6}{c|}{1-shot}&\multicolumn{6}{c|}{5-shot}\\
\cline{3-14}
& &\multicolumn{5}{c|}{\textit{mIoU}}&{\textit{FB-IoU}}&\multicolumn{5}{c|}{\textit{mIoU}}&{\textit{FB-IoU}}\\\cline{3-14}
& & fold1 & fold2 & fold3 & fold4 & mean & mean & fold1 & fold2 & fold3 & fold4 & mean & mean\\
\hline
\multirow{4}{*}{\rotatebox[origin=c]{90}{Baselines}}
& PAPNet~\cite{liu2020part} & 52.7 & 62.8 & 57.4 & 47.7 & 55.2 & - & 60.3 & 70.0 & \textbf{69.4} & \textbf{60.7} & 65.1 & -\\
& DAN~\cite{yang2020new} & 54.7 & 68.6 & 57.8 & 51.6 & 58.2 & 71.9 & 57.9 & 69.0 & 60.1 & 54.9 & 60.5 & 72.3\\
& DENet~\cite{liu2020dynamic} & 55.7 & 69.7 & 63.2 & 51.3 & 60.1 & - & 54.7 & 71.0 & 64.5 & 51.6 & 60.5 & -\\
& PFENet~\cite{tian2020prior} & 61.7 & 69.5 & 55.4 & 56.3 & 60.8 & 73.3 & 63.1 & 70.7 & 55.8 & 57.9 & 61.9 & 73.9\\
\hline
\multirow{4}{*}{\rotatebox[origin=c]{90}{Ours}}
& Ours & \textbf{62.4} & \textbf{74.3} & \textbf{67.0} & \textbf{58.4} & \textbf{65.5} & \textbf{76.7} & \textbf{63.5} & \textbf{74.1} & 67.3 & 58.9 & \textbf{65.9} & \textbf{76.9}\\
& Ours-noObj & 60.9 & 71.7 & 64.0 & 55.6 & 63.0 & 74.4 & 62.9 & 72.1 & 64.3 & 57.1 & 64.1 & 75.4\\
& Ours-DENet & 57.7 & 72.6 & 66.3 & 54.4 & 62.7 & 75.8 & 57.5 & 72.1 & 69.2 & 54.8 & 63.4 & 76.0\\
\cline{2-14}
& Ours-gtObj & 84.2 & 80.3 & 74.5 & 82.1 & 80.3 & 86.4 &  88.3 & 81.2 & 76.4 & 83.2 & 82.3 & 88.4\\
\hline
\end{tabular}
\end{center}
\vspace{-6pt}
\caption{Performance of the top baselines and our approach on 1-shot and 5-shot semantic segmentation for PASCAL-$5^i$. Ours-gtObj shows the upper bound by using ground truth objectness. (``-" means the original paper does not report its performance for this metric.)}
\label{table:pascal_1shot}
\end{table*}

\subsection{Overall Performance: PASCAL-$5^i$}
\label{sec:experiment_pascal}
\paragraph{Dataset:} PASCAL-$5^i$~\cite{shaban2017one} is a dataset for few-shot semantic segmentation, built from PASCAL VOC 2012~\cite{everingham2015pascal} with annotations augmented to it~\cite{hariharan2014simultaneous}. We follow the dataset split in \cite{shaban2017one} such that 20 object categories are evenly divided into four folds, each with five categories. The category split in each fold is as follows---\textbf{fold1}: aeroplane, bicycle, bird, boat, bottle; \textbf{fold2}: bus, car, cat, chair, cow; \textbf{fold3}: dining table, dog, horse, motorbike, person; \textbf{fold4}: potted plant, sheep, sofa, train, tv/monitor. 

We perform cross-validation, employing three folds for training and the remaining fold for evaluation.  Consequently, when training both the objectness module and the comparison module, our model never observes any object from the testing categories. At test time, we randomly sample 1,000 episodes (i.e., a query image $\mathcal{Q}$ and a set $\mathcal{S}$ of support images) in the test fold for evaluation. 

\vspace{-1em}\paragraph{Our Approach:} 
 We evaluate four variants.
 
\textit{Ours}: This is model described in Section~\ref{sec:framework}: i.e., it uses HRNetV2-W48~\cite{sun2019high} as the backbone of the objectness module, ResNet-50~\cite{he2016deep} as the backbone of the feature extraction module, and FEM to produce the multi-scale representation in the multi-scale dense comparison module.


\textit{Ours-noObj}: An ablated variant of \textit{Ours} that lacks the objectness module to help reveal the benefit of objectness.

\textit{Ours-DENet}: This variant demonstrates the benefit of objectness for different base architectures, which is particularly relevant given that few shot segmentation is a hot field with new methods released rapidly.\footnote{https://github.com/xiaomengyc/Few-Shot-Semantic-Segmentation-Papers}  This variant augments our objectness module to the code base for the current state-of-art method, DENet~\cite{liu2020dynamic}, by concatenating an additional channel of the objectness map produced by our objectness module with the query features extracted by DENet.

\textit{Ours-gtObj}: This variant demonstrates the upper bound performance of \textit{Ours} with the ground truth objectness.

\vspace{-1em}\paragraph{Baselines:} 
We compare our approaches to 19 few-shot segmentation methods: OSLSM~\cite{shaban2017one}, co-FCN~\cite{rakelly2018conditional}, PL~\cite{dong2018few}, A-MCG~\cite{hu2019attention}, AMP~\cite{siam2019amp}, SG-One~\cite{zhang2018sg}, PANet~\cite{wang2019panet}, CANet~\cite{zhang2019canet}, PGNet~\cite{zhang2019pyramid}, FWB~\cite{nguyen2019feature} ,LTM~\cite{yang2020new}, PAPNet~\cite{liu2020part}, CRNet~\cite{liu2020crnet}, PMM~\cite{yang2020prototype}, SPNet~\cite{gairola2020simpropnet}, SST~\cite{zhu2020self}, DAN~\cite{yang2020new}, DENet~\cite{liu2020dynamic} and PFENet~\cite{tian2020prior}. We report numbers from their original papers.  Due to space constraints, we report results for the weakest 15 baselines in the Supplementary Materials.  Given that existing methods employ different data loaders during training, to support fair comparison, we employ data loaders of two state-of-art methods that we aim to demonstrate advantages over: PFENet (for \emph{Ours} and \emph{Ours-noObj}) and DENet (for \emph{Ours-DeNet}). 


\begin{figure*}[!th]
\centering
\includegraphics[width=0.98\textwidth]{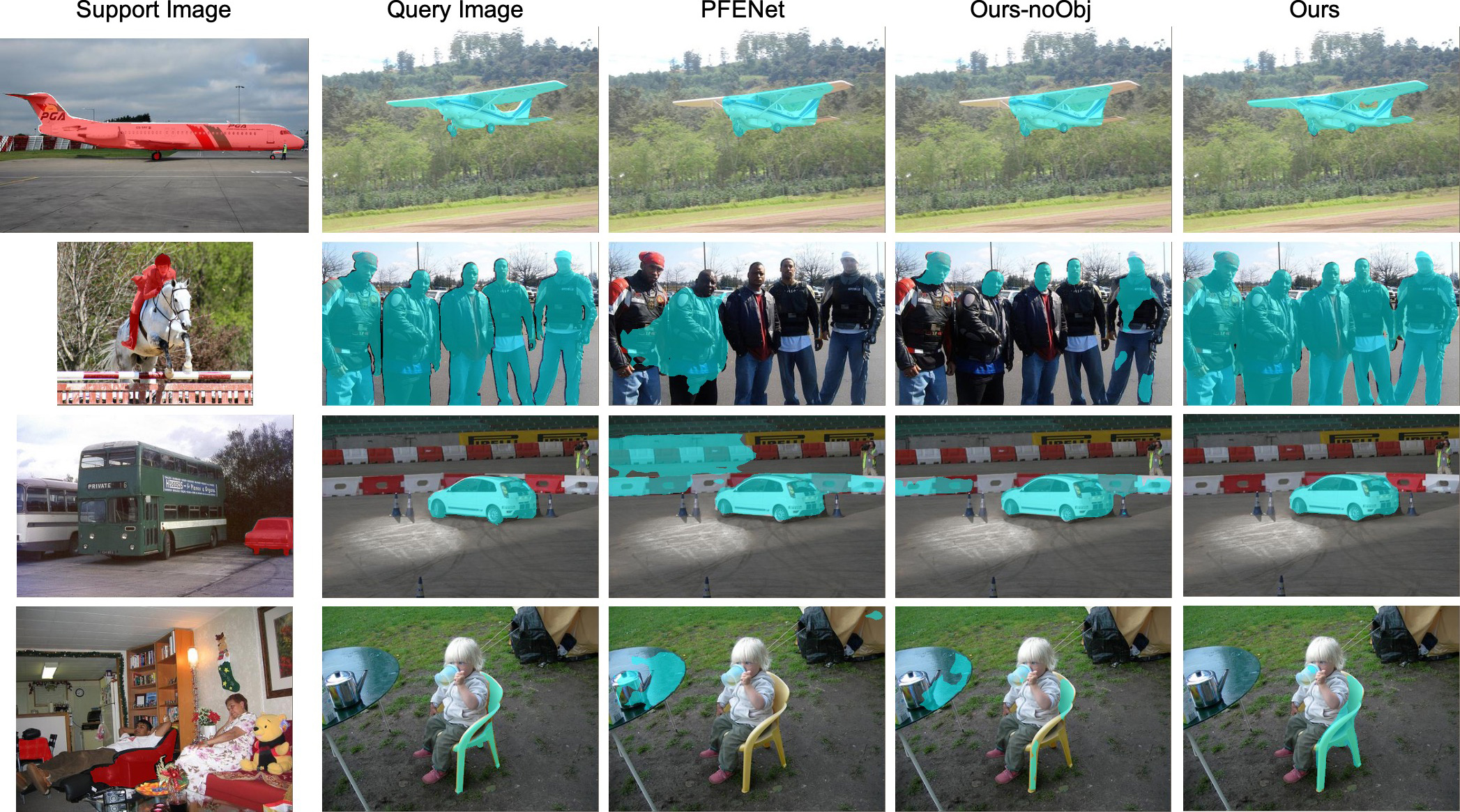}
\caption{Qualitative results for 1-shot semantic segmentation on PASCAL-$5^i$. From left to right we show the support image with the mask of the target category, the query image with ground truth segmentation for the target category, the prediction of PFENet~\cite{tian2020prior}, the prediction of our implemented baseline without objectness and the prediction of our method with objectness.}
\label{fig:qualitative_pascal}
\end{figure*}

\vspace{-1em}\paragraph{Evaluation Metrics:} 
We use two popular evaluation metrics: mean Intersection-over-Union (\textit{mIoU}) and Foreground-Background Intersection-over-Union (\textit{FB-IoU}). \textit{mIoU} \cite{shaban2017one,zhang2018sg} measures per-class foreground IoU and averages the foreground IoU over all classes. \textit{FB-IoU}~\cite{dong2018few,rakelly2018conditional,hu2019attention} discards category information and treats all categories as foreground. Then it measures the mean of foreground IoU and background IoU over all test images. 

\vspace{-1em}\paragraph{Few-Shot Semantic Segmentation Results:} 
Results for 1-shot and 5-shot segmentation are shown in Table~\ref{table:pascal_1shot}. 

We observe that incorporating objectness consistently results in considerable gains over ablated variants that lack objectness.  For instance, \textit{Ours} outperforms \textit{Ours-noObj} by 2.5 and 2.3 percentage points in \textit{mIoU} and \textit{FB-IoU} for 1-shot segmentation, and 1.8 and 1.5 percentage points in \textit{mIoU} and \textit{FB-IoU} for 5-shot segmentation.  Similarly, \textit{Ours-DENet} outperforms DENet~\cite{liu2020dynamic} by 2.6 and 2.9 percentage points in \textit{mIoU} for 1-shot and 5-shot segmentation respectively. Altogether, these findings highlight that our approach to increase model capacity using objectness to achieve improved segmentation, successfully generalizes to different frameworks.

Moreover, we observe that our model (\textit{Ours}) achieves state-of-art performance with respect to both metrics for both 1-shot and 5-shot segmentation.{\footnote{We found that this performance gain occurs only when we train the modules independently.  When training the modules jointly, we observe \emph{worse} results; i.e., a 11.7\% drop in mIoU and 7.2\% in FB-IoU compared to \textit{Ours} in 1-shot. In other words, the training scheme is a key ingredient to successfully increase our model’s capacity without severe overfitting.}$^,$\footnote{We validate the gain comes from introducing class-agnostic objectness rather than ensembling of two networks or conducting an auxiliary task in the Supplementary Materials.}}  For instance, in 1-shot segmentation \textit{Ours} outperforms the next best baseline (PFENet) by 4.7 and 3.4 percentage points in \textit{mIoU} and \textit{FB-IoU} respectively. In 5-shot, it outperforms all baselines by at least 0.8 and 3.0 percentage points in \textit{mIoU} and \textit{FB-IoU} respectively. The gain of our 5-shot results over 1-shot shows our method benefits from more support information.

There is still a performance gap between \textit{Ours} and \textit{Ours-gtObj}, the upper bound when using ground truth objectness; e.g. \textit{Ours-gtObj} outperforms \textit{Ours} by 14.8 and 16.4 percentage points in terms of \textit{mIoU} in 1-shot and 5-shot results. The gap indicates the error introduced by our objectness module and highlights the potential room for improvement of our method. 

We show qualitative results in Figure~\ref{fig:qualitative_pascal}. Qualitatively our method (column 5) produces more accurate segmentations than PFENet~\cite{tian2020prior} (column 3) and our implementation without objectness (column 4). In rows 1 and 2, both PFENet~\cite{tian2020prior} and \textit{Ours-noObj} only correctly segment part of the target region due to significant appearance variance between the support and query image, while \textit{Ours} produces an accurate segmentation by leveraging objectness. In rows 3 and 4, both PFENet~\cite{tian2020prior} and \textit{Ours-noObj} misclassify some background regions as the target object category, while \textit{Ours} correctly labels them as negative by taking advantage of objectness.

More generally, from inspection of resulting segmentations, we believe the increased capacity from objectness enables our method to augment more effective localization of object boundary.  We illustrate this intuition with additional qualitative figures in the Supplemental Materials (i.e., Figure 5).  Specifically, the previous state-of-art method, PFENet, \emph{can} identify the correct object in all examples but \emph{cannot} produce good segmentation boundaries like our method.   These qualitative examples also highlight our method's advantage for when the query and support images both represent a single and multiple target objects.

\subsection{Overall Performance: COCO-$20^i$}
\label{sec:experiment_coco}

\begin{table*}[!t]
\begin{center}
\begin{tabular}{| c  l | c  c  c  c | c | c | c  c  c  c | c | c |}
\hline
&\multirow{3}{*}{Method}&\multicolumn{6}{c|}{1-shot}&\multicolumn{6}{c|}{5-shot}\\
\cline{3-14}
& &\multicolumn{5}{c|}{\textit{mIoU}}&{\textit{FB-IoU}}&\multicolumn{5}{c|}{\textit{mIoU}}&{\textit{FB-IoU}}\\\cline{3-14}
& & fold1 & fold2 & fold3 & fold4 & mean & mean & fold1 & fold2 & fold3 & fold4 & mean & mean\\
\hline
\multirow{4}{*}{\rotatebox[origin=c]{90}{Baselines}} 
& PAPNet~\cite{liu2020part} & 36.5 & 26.5 & 26.0 & 19.7 & 27.2 & - & \textbf{48.9} & 31.4 & 36.0 & 30.6 & 36.7 & -\\
& PMM~\cite{yang2020prototype} & 29.5 & 36.8 & 28.9 & 27.0 & 30.6 & - & 33.8 & 42.0 & 33.0 & 33.3 & 35.5 & -\\
& PFENet~\cite{tian2020prior} & 33.4 & 36.0 & 34.1 & 32.8 & 34.1 & 60.0 & 35.9 & 40.7 & 38.1 & 36.1 & 37.7 & 61.6\\
& DENet~\cite{liu2020dynamic} & 42.9 & 45.8 & \textbf{42.2} & 40.2 & 42.8 & - & 45.4 & 44.9 & 41.6 & 40.3 & 43.0 & -\\
\hline
\multirow{4}{*}{\rotatebox[origin=c]{90}{Ours}} 
& Ours & 35.0 & 42.7 & 38.1 & 37.9 & 38.4 & 66.9 & 37.0 & 46.4 & 41.1 & 41.3 & 41.5 & \textbf{68.7}\\
& Ours-noObj & 34.1 & 42.3 & 36.8 & 35.0 & 37.0 & 64.7 & 35.4 & \textbf{47.8} & 40.6 & 37.9 & 40.4 & 66.4\\
& Ours-DENet & \textbf{48.0} & \textbf{46.9} & 41.6 & \textbf{40.8} & \textbf{44.3} & \textbf{67.7} & 48.8 & 47.5 & \textbf{43.7} & \textbf{44.5} & \textbf{46.1} & 68.3\\
\cline{2-14}
& Ours-gtObj & 44.8 & 56.6 & 53.9 & 50.2 & 51.4 & 74.4 & 48.4 & 63.6 & 57.5 & 55.0 & 56.1 & 77.4\\

\hline
\end{tabular}
\end{center}
\vspace{-6pt}
\caption{Performance of the top baselines and our approach on 1-shot and 5-shot semantic segmentation for COCO-$20^i$. Ours-gtObj shows the upper bound by using ground truth objectness. (``-" means the original paper does not report performance for this metric.)}
\label{table:coco_1shot}
\end{table*}

\paragraph{Dataset:} 
COCO-$20^i$~\cite{wang2019panet,nguyen2019feature}, created from MSCOCO~\cite{lin2014microsoft}, is a more challenging dataset than PASCAL-$5^i$. Its 80 object categories are evenly divided into four folds, with each fold containing 20 categories. We follow the same splits as \cite{tian2020prior}. The set of class indexes contained in fold $i$ is written as $\{4j-3+i\}$ where $j\in\{1,2,\dots,20\}$, $i\in\{0,1,2,3\}$. Because there are many more images in COCO than PASCAL, as suggested by prior work~\cite{tian2020prior}, we sample 20,000 instead of 1,000 episodes to produce reliable and stable results.


\vspace{-1em}\paragraph{Our Approach:} As in Section~\ref{sec:experiment_pascal}, we evaluate four variants of our approach: \textit{Ours}, \textit{Ours-noObj}, \textit{Ours-DENet} and \textit{Ours-gtObj}.

\vspace{-1em}\paragraph{Baselines:} We compare our approach to eight methods: PANet~\cite{wang2019panet}, FWB~\cite{nguyen2019feature}, PAPNet~\cite{liu2020part},  PMM~\cite{yang2020prototype},  SST~\cite{zhu2020self}, DAN~\cite{yang2020new}, PFENet~\cite{tian2020prior} and DENet~\cite{liu2020dynamic}.\footnote{This number is smaller than for the previous dataset because some methods reported in Table~\ref{table:pascal_1shot} did not share results for this dataset.}  Results for the weakest baselines are in the Supplementary Materials. 

\vspace{-1em}\paragraph{Evaluation Metrics:} As for Section~\ref{sec:experiment_pascal}, we again use \textit{mIoU} and \textit{FB-IoU}.

\vspace{-1em}\paragraph{Few-Shot Semantic Segmentation Results:} 
We report 1-shot and 5-shot segmentation results in Table~\ref{table:coco_1shot}. 

We again observe that incorporating objectness consistently results in considerable gains over ablated variants that lack objectness.  For instance, \textit{Ours} outperforms \textit{Ours-noObj} in 1-shot with gains of 1.4 and 2.2 percentage points in \textit{mIoU} and \textit{FB-IoU} respectively. Additionally, \textit{Ours-DENet} outperforms DENet~\cite{liu2020dynamic} by 1.5 and 3.1 percentage points in \textit{mIoU} for 1-shot and 5-shot segmentation respectively. These ablated variants reinforce that our approach for increasing model capacity by introducing objectness is beneficial despite that greater capacity typically leads to overfitting in the few shot setting. 

We observe state-of-art performance with respect to \textit{mIoU} in both 1-shot and 5-shot results from our DENet variant that uses objectness: \textit{Ours-DENet}.\footnote{\textit{Ours-DENet} outperforms \textit{Ours} because DENet uses a different mechanism to encode class-specific features, which we hypothesize is better at dealing with the greater diversity of categories in COCO.} In 1-shot, our method outperforms all methods by at least 1.5 and 5.4 percentage points in \textit{mIoU} and \textit{FB-IoU} respectively. In 5-shot, our approach outperforms all methods by at least 3.1 and 4.8 percentage points in \textit{mIoU} and \textit{FB-IoU} respectively.  As before, we observe a better performance of our 5-shot results over 1-shot results, reinforcing that our method benefits from more support information. The gap between \textit{Ours} and \textit{Ours-gtObj} again indicates the noise introduced by our objectness module and highlights the potential of our method. 

Compared to results on PASCAL-$5^i$, the accuracy is considerably lower on COCO-$20^i$. A regular reason for failure is the comparison module inaccurately segments the target object from the regions produced by the objectness module (this error is exemplified in Figure~\ref{fig:qualitative_pascal} where part of the cup in row 4 also appears in our final prediction). Class-specific features learned by the comparison module may be prone to overfitting for the greater diversity of categories in COCO. 

\subsection{Fine-Grained Analysis of Design Choices}
\label{sec:ablation}

\paragraph{Our Approaches.}
We examine three types of design choices, as summarized below.

\emph{Feature extraction module training:} We fine-tune the feature extraction module in our base model (\emph{Ours}) rather than freeze it in training and leave out the objectness module.  This variant motivates the premise of this paper that class-specific features are prone to overfitting.

\emph{Feature extraction module}: We evaluate three different backbone models for the feature extraction module (which remains fixed during training to mitigate overfitting): ResNet-50, ResNet-101~\cite{he2016deep} and HRNetV2-W48.  We evaluate three implementations which uses HRNetV2-48~\cite{sun2019high} as the backbone of the objectness module (which is not fixed during training) and an additional three implementations that lack an objectness module.  Altogether, these six variants support examining the benefit of different feature spaces (but not the model capacity, since all studied models have the same model capacity).  

\emph{Objectness module}: Next, we investigate the influence of different backbone models for the objectness module which, in turn, drives the overall model capacity.  To do so, we hold the model's feature extractor as ResNet-50 while varying the objectness backbone to use one of the following: ResNet-50, ResNet-101~\cite{he2016deep} and HRNetV2-W48~\cite{sun2019high}.  As a lower bound, we also evaluate a fourth implementation where this objectness module is removed.  

\vspace{-1em}\paragraph{Experimental Set-up.}
Observing that training time for our model requires more than 80 hours for a single fold with Nvidia Tesla V100, we focus on a more practical setting.  We simplify training image pre-processing by employing a data loader that does {\em not} perform training data augmentation and {\em does} resize all images to a smaller resolution of $328\times328$.  We also adopt \textit{ASPP}~\cite{chen2017rethinking} as the multi-scale dense comparison module.  We conduct our experiments on PASCAL-$5^i$  for 1-shot learning, and use as evaluation metrics \textit{mIoU} and \textit{FB-IoU}.

\setlength{\tabcolsep}{4pt}
\begin{table}[!bt]
\begin{center}
\begin{tabular}{ | l | c  c  c  c | c | c |}
\hline
\multirow{2}{*}{Feature}&\multicolumn{5}{c|}{\textit{mIoU}}&{\textit{FB-IoU}}\\\cline{2-7}
& fold1 & fold2 & fold3 & fold4 & mean & mean\\
\hline
Res50* & 48.9 & 63.8 & 48.3 & 45.9 & 51.7 & 70.2\\
Res101* & 50.1 & 63.0 & 47.9 & 45.7 & 51.7 & 70.7\\
HRNet* & 53.3 & 63.6 & 46.0 & 43.1 & 51.5 & 70.9\\
\hline
HRNet*-F &50.6 & 61.1 & 46.2 & 41.2 & 49.8 & 70.1\\
\hline
Res50 & 56.9 & 66.4 & 57.1 & 50.7 & 57.8 & 73.0\\
Res101 & 57.0 & 65.9 & \textbf{58.5} & 50.3 & 57.9 & 73.4\\
HRNet & \textbf{61.2} & \textbf{67.7} & 56.5 & \textbf{52.5} & \textbf{59.5} & \textbf{73.9}\\
\hline
\end{tabular}
\end{center}
\vspace{-6pt}
\caption{\textit{mIoU} and \textit{FB-IoU} with multiple backbones as the feature extraction module for 1-shot segmentation on PASCAL-$5^i$. Results without objectness are in rows 1-4. Weights of the feature extraction module are frozen in training for rows 1-3, while fine-tuned for row 4. Rows 5-7 show results with HRNetV2-48 as the backbone of the objectness module while freezing the feature extraction module in training.}
\vspace{-1pt}
\label{table:pascal_1shot_feat}
\end{table}

\vspace{-1em}\paragraph{Results.} We evaluate three types of design choices. 

\emph{Feature extraction module training:} Results comparing the two training strategies for the feature extraction module---fine-tuning and freezing---are shown in Table~\ref{table:pascal_1shot_feat}. We observe that finetuning the feature extraction module (row 4) results in worse performance than freezing it (row 3) in training; e.g. \textit{mIoU} and \textit{FB-IoU} decrease by 1.7 and 0.8 percentage point respectively. This reinforces the findings of prior work~\cite{tian2020prior} that training with all the backbone parameters causes significant performance reduction, likely due to overfitting to the training classes. 

\setlength{\tabcolsep}{4pt}
\begin{table}[!bt]
\begin{center}
\begin{tabular}{ | l | c  c  c  c | c | c |}
\hline
\multirow{2}{*}{Objectness}&\multicolumn{5}{c|}{\textit{mIoU}}&{\textit{FB-IoU}}\\\cline{2-7}
& fold1 & fold2 & fold3 & fold4 & mean & mean\\
\hline
- - & 48.9 & 63.8 & 48.3 & 45.9 & 51.7 & 70.2\\
Res50 & 49.4 & 63.7 & 56.4 & 44.3 & 53.4 & 70.7\\
Res101 & 50.6 & 64.5 & 55.3 & 46.6 & 54.2 & 71.0\\
HRNet & \textbf{56.9} & \textbf{66.4} & \textbf{57.1} & \textbf{50.7} & \textbf{57.8} & \textbf{73.0}\\
\hline
\end{tabular}
\end{center}
\vspace{-6pt}
\caption{\textit{mIoU} and \textit{FB-IoU} of our approach with ResNet-50 as the feature extractor and multiple backbones as the objectness module on the task of 1-shot segmentation in PASCAL-$5^i$. The first row shows the results of a baseline without an objectness module.}
\vspace{-1pt}
\label{table:pascal_1shot_obj}
\end{table}

\emph{Feature extraction module}: Results from using different backbone models for the feature extraction module are shown in Table~\ref{table:pascal_1shot_feat}. 

As shown in rows 5-7 of Table~\ref{table:pascal_1shot_feat}, HRNetV2-W48 projects images into a better feature space than ResNet-50 and ResNet-101 when leveraging objectness, i.e. HRNetV2-W48 improves \textit{mIoU} and \textit{FB-IoU} by at least 1.6 and 0.5 percentage points respectively compared to ResNet-50 and ResNet-101. Recall, we freeze the feature extraction module in training. Thus, we are evaluating the feature space provided by this module, rather than its capacity. 

When lacking objectness (Table~\ref{table:pascal_1shot_feat} row 1-3), we observe similar \textit{mIoU} for the three backbones, but \textit{FB-IoU} increases from ResNet-50 to ResNet-101 and HRNetV2-W48. Table~\ref{table:pascal_1shot_feat} also shows consistent gains when pairing our objectness module with different backbones as the feature extraction module; e.g., at least a gain of 6.1 and 2.7 percentage points in \textit{mIoU} and \textit{FB-IoU} respectively.

\emph{Objectness module}: Results for using backbones of different capacity for the objectness module are reported in Table~\ref{table:pascal_1shot_obj}. Compared to not using an objectness module (row 1), we observe gains in both \textit{mIoU} and \textit{FB-IoU} when introducing the objectness module (row 2,3,4). The gain grows as the objectness module's capacity increases; e.g., the gain in \textit{mIoU} and \textit{FB-IoU} boost from 1.7 to 6.1 and from 0.5 to 2.8 percentage points respectively, when we use HRNetV2-W48 (row 4) instead of ResNet-50 (row 2). These findings demonstrate both an advantage of introducing the objectness module and increasing its capacity.\footnote{The baseline without objectness (row 1 of Table~\ref{table:pascal_1shot_obj}) differs from the performance of \textit{Ours-noObj} in Table~\ref{table:pascal_1shot} because different data pre-processing pipelines and objectness-aware dense comparison modules are used (to achieve a more practical training time for fine-grained analysis).} 
\section{Conclusions}
We demonstrate that increasing model capacity can achieve improved performance for few-shot semantic segmentation when introducing a class-agnostic objectness module. Experiments highlight key design choices that are necessary for successfully leveraging objectness. Experiments also demonstrate our method's success generalizes across different datasets, base models, and backbone architectures. Our approach avoids the severe overfitting that is typical when increasing model capacity in the presence of little training data.  Our findings offer promising evidence that objectness can benefit new methods that rely on existing or future backbone architectures, data loaders, and training schedules.



{\small
\bibliographystyle{ieee_fullname}
\bibliography{egbib}
}

\clearpage
\onecolumn
\newpage
\appendix
\noindent {\LARGE \textbf{Appendix}}
\vspace{2em}

This document supplements Section 4 of the main paper.  In particular, it includes the following:

\begin{itemize}
\item Implementation details for the four variants of our method (supplements \textbf{Section 4.1, Section 4.2})
\item Quantitative results of our method and all the 19 baselines in 1-shot and 5-shot semantic segmentation on PASCAL-$5^i$ (supplements \textbf{Section 4.1, Table 1})
\item Quantitative results of our method and all the eight baselines in 1-shot and 5-shot semantic segmentation on COCO-$20^i$ (supplements \textbf{Section 4.2, Table 2})
\item Qualitative examples demonstrating success and failure cases of our method on PASCAL-$5^i$ (supplements \textbf{Section 4.1}) 
\item Qualitative examples demonstrating success and failure cases of our approach on COCO-$20^i$ (supplements \textbf{Section 4.2})
\item Parallel fine-grained analysis of the \textit{feature extraction module} on COCO-$20^i$ (supplements \textbf{Section 4.3})
\end{itemize}

\section{Implementation Details for Four Variants of Our Method (Sections 4.1 and 4.2)}
We observe that existing few-shot semantic segmentation methods employ different data loaders and training schedules. In our experiments, we find that they can significantly affect the performance. To support fair comparison in Sections 4.1 and 4.2 of the main paper, we employ data loaders and training schedules of the top two performing methods that we aim to demonstrate advantages over: PFENet~\cite{tian2020prior} (for \emph{Ours}, \emph{Ours-noObj} and \emph{Ours-gtObj}) and DENet~\cite{liu2020dynamic} (for \emph{Ours-DeNet}). We elaborate below on the data pre-processing steps and training details employed for the four variants of our method (\emph{Ours}, \emph{Ours-noObj}, \emph{Ours-DeNet} and \emph{Ours-gtObj}). 

PFENet (for \emph{Ours}, \emph{Ours-noObj} and \emph{Ours-gtObj}): During training, images are processed by random horizontally flipping, random scaled between a given range\footnote{$[0.9, 1.1]$ for PASCAL-$5^i$ and $[0.8, 1.25]$ for COCO-$20^i$.}, random rotation from -10 to 10 degrees and random Gaussian blur, as that of \cite{tian2020prior}. Finally patches are randomly cropped from the processed images as the training samples.\footnote{$473\times473$ for PASCAL-$5^i$ and $641\times641$ for COCO-$20^i$.} At inference time, each image is also resized to the randomly cropped patch size while keeping the original aspect ratio by padding zeros. The prediction result is resized back to the original label size for evaluation. As done in PFENet~\cite{tian2020prior,zhang2019canet}, our model is trained for 50 epochs, with learning rate 0.005 and batch size 8 on PASCAL-$5^i$. We train our model for 200 epochs with learning rate 0.0025 and batch size 4 on COCO-$20^i$.

DENet (for \emph{Ours-DENet}): During training, unlike PFENet~\cite{tian2020prior}, only random horizontal flipping is used for data augmentation. The images used for training are resized to $321\times321$ for both PASCAL-$5^i$ and COCO-$20^i$, rather than randomly cropped from the processing images. At inference time, each image is also resized to $321\times321$ for both PASCAL-$5^i$ and COCO-$20^i$. As done in DENet~\cite{liu2020dynamic}, \textit{Ours-DENet} is trained for 150k iterations, with learning rate 0.0025 and batch size 8 on PASCAL-$5^i$. We train \textit{Ours-DENet} for 200k iterations with learning rate 0.0025 and batch size 8 on COCO-$20^i$.

\section{Quantitative Results in PASCAL-$5^i$}

Due to space constraints we only report results for the best-performing four baselines in the main paper. We report here results for all 19 baselines and our method in Table~\ref{table:pascal_1shot_supp}. 

We observe that incorporating objectness consistently results in considerable gains over ablated variants that lack objectness.  For instance, \textit{Ours} outperforms \textit{Ours-noObj} by 2.5 and 2.3 percentage points in \textit{mIoU} and \textit{FB-IoU} for 1-shot segmentation, and 1.8 and 1.5 percentage points in \textit{mIoU} and \textit{FB-IoU} for 5-shot segmentation.  Similarly, \textit{Ours-DENet} outperforms DENet~\cite{liu2020dynamic} by 2.6 and 2.9 percentage points in \textit{mIoU} for 1-shot and 5-shot segmentation respectively. Altogether, these findings highlight that our approach to increase model capacity using objectness to achieve improved segmentation successfully generalizes to different frameworks.

Moreover, we observe that our model (\textit{Ours}) achieves state-of-art performance with respect to both metrics for both 1-shot and 5-shot segmentation. For instance, in 1-shot segmentation our method (\textit{Ours}) outperforms the next best baseline (PFENet) by 4.7 and 3.4 percentage points in \textit{mIoU} and \textit{FB-IoU} respectively. In 5-shot, our approach (\textit{Ours}) outperforms all baselines by at least 0.8 and 2.3 percentage points in \textit{mIoU} and \textit{FB-IoU} respectively. 

We conduct ablation studies to further validate the gain comes from introducing class-agnostic objectness rather than the ensembling of two networks. We evaluate another variant of our architecture without objectness which ensembles two backbones (ResNet-50 and HRNetV2-W48) inside the feature extraction module from Section 3.2.2. We fix both backbones in training since finetuning yields worse results. We observe \emph{worse} results; e.g., 1.4 and 3.2 percentage points drop in mIoU and FB-IoU respectively in terms of 1-shot segmentation compared to \textit{Ours}. 

We also conduct ablation studies to validate the gain does not come from conducting an auxiliary task. We train \textit{Ours-noObj} with an auxiliary task of predicting the objectness mask for one joint class of all semantic classes (called \textit{Ours-noObj-aux}); all parameters are trained jointly (including the feature extraction module). We observe a considerable performance drop: \textit{Ours-noObj-aux} achieves 52.4 in \textit{mIoU} in terms of 1-shot segmentation, which is 13.1 and 10.6 percentage points lower than \textit{Ours} and \textit{Ours-noObj}. Therefore, the gain of \textit{Ours} is not from conducting an auxiliary task.

\begin{table}[!bt]
\begin{center}
\begin{tabular}{| l | c  c  c  c | c | c | c  c  c  c | c | c |}
\hline
\multirow{3}{*}{Method}&\multicolumn{6}{c|}{1-shot}&\multicolumn{6}{c|}{5-shot}\\
\cline{2-13}
&\multicolumn{5}{c|}{\textit{mIoU}}&{\textit{FB-IoU}}&\multicolumn{5}{c|}{\textit{mIoU}}&{\textit{FB-IoU}}\\\cline{2-13}
& fold1 & fold2 & fold3 & fold4 & mean & mean & fold1 & fold2 & fold3 & fold4 & mean & mean\\
\hline
OSLSM~\cite{shaban2017one} & 33.6 & 55.3 & 40.9 & 33.5 & 40.8 & 61.3 & 35.9 & 58.1 & 42.7 & 39.1 & 43.9 & 61.5\\
co-FCN~\cite{rakelly2018conditional} & 36.7 & 50.6 & 44.9 & 32.4 & 41.1 & 60.1 & 37.5 & 50.0 & 44.1 & 33.9 & 41.4 & 60.2\\
PL~\cite{dong2018few} & - & - & - & - & - & 61.2 & - & - & - & - & - & 62.3\\
A-MCG~\cite{hu2019attention} & - & - & - & - & - & 61.2 & - & - & - & - & - & 62.2\\
AMP~\cite{siam2019amp} & 41.9 & 50.2 & 46.7 & 34.7 & 43.4 & 62.2 & 41.8 & 55.5 & 50.3 & 39.9 & 46.9 & 63.8\\
SG-One~\cite{zhang2018sg} & 40.2 & 58.4 & 48.4 & 38.4 & 46.3 & 63.1 & 41.9 & 58.6 & 48.6 & 39.4 & 47.1 & 65.9\\
PANet~\cite{wang2019panet} & 42.3 & 58.0 & 51.1 & 41.3 & 48.1 & 66.5 & 51.8 & 64.6 & 59.8 & 46.5 & 55.7 & 70.7\\
CANet~\cite{zhang2019canet} & 52.5 & 65.9 & 51.3 & 51.9 & 55.4 & 66.2 & 55.5 & 67.8 & 51.9 & 53.2 & 57.1 & 69.6\\
PGNet~\cite{zhang2019pyramid} & 56.0 & 66.9 & 50.6 & 50.4 & 56.0 & 69.9 & 57.7 & 68.7 & 52.9 & 54.6 & 58.5 & 70.5\\
FWB~\cite{nguyen2019feature} & 51.3 & 64.5 & 56.7 & 52.2 & 56.2 & - & 54.8 & 67.4 & 62.2 & 55.3 & 59.9 & -\\
LTM~\cite{yang2020new} & 52.8 & 69.6 & 53.2 & 52.3 & 57.0 & 71.8 & 57.9 & 69.9 & 56.9 & 57.5 & 60.6 & 74.6\\
PAPNet~\cite{liu2020part} & 52.7 & 62.8 & 57.4 & 47.7 & 55.2 & - & 60.3 & 70.0 & \textbf{69.4} & \textbf{60.7} & 65.1 & -\\
CRNet~\cite{liu2020crnet} & - & - & - & - & 55.7 & 66.8 & - & - & - & - & 58.8 & 71.5\\
PMM~\cite{yang2020prototype} & 55.2 & 66.9 & 52.6 & 50.7 & 56.3 & - & 56.3 & 67.3 & 54.5 & 51.0 & 57.3 & -\\
SPNet~\cite{gairola2020simpropnet} & 54.9 & 67.3 & 54.5 & 52.0 & 57.2 & - & 57.2 & 68.5 & 58.4 & 56.1 & 60.0 & -\\
SST~\cite{zhu2020self} & 54.4 & 66.4 & 57.1 & 52.5 & 57.6 & - & 58.6 & 68.7 & 63.1 & 55.3 & 61.4 & -\\
DAN~\cite{yang2020new} & 54.7 & 68.6 & 57.8 & 51.6 & 58.2 & 71.9 & 57.9 & 69.0 & 60.1 & 54.9 & 60.5 & 72.3\\
DENet~\cite{liu2020dynamic} & 55.7 & 69.7 & 63.2 & 51.3 & 60.1 & - & 54.7 & 71.0 & 64.5 & 51.6 & 60.5 & -\\
PFENet~\cite{tian2020prior} & 61.7 & 69.5 & 55.4 & 56.3 & 60.8 & 73.3 & 63.1 & 70.7 & 55.8 & 57.9 & 61.9 & 73.9\\
\hline
Ours & \textbf{62.4} & \textbf{74.3} & \textbf{67.0} & \textbf{58.4} & \textbf{65.5} & \textbf{76.7} & \textbf{63.5} & \textbf{74.1} & 67.3 & 58.9 & \textbf{65.9} & \textbf{76.9}\\
Ours-noObj & 60.9 & 71.7 & 64.0 & 55.6 & 63.0 & 74.4 & 62.9 & 72.1 & 64.3 & 57.1 & 64.1 & 75.4\\
Ours-DENet & 57.7 & 72.6 & 66.3 & 54.4 & 62.7 & 75.8 & 57.5 & 72.1 & 69.2 & 54.8 & 63.4 & 76.0\\
\cline{1-13}
Ours-gtObj & 84.2 & 80.3 & 74.5 & 82.1 & 80.3 & 86.4 &  88.3 & 81.2 & 76.4 & 83.2 & 82.3 & 88.4\\
\hline
\end{tabular}
\end{center}
\vspace{-0pt}
\caption{\textit{mIoU} and \textit{FB-IoU} of 19 baselines and our approach on the task of 1-shot and 5-shot semantic segmentation for PASCAL-$5^i$. Ours-gtObj shows the upper bound by using ground truth objectness. (“-” means the original paper does not report performance for this metric.)}
\vspace{0pt}
\label{table:pascal_1shot_supp}
\end{table}

\section{Quantitative Analysis in COCO-$20^i$}

Due to space constraints we only report results for the best-performing four baselines in the main paper.  We report here results for all eight baselines and our method in Table~\ref{table:coco_1shot_supp}.

 We observe that incorporating objectness consistently results in considerable gains over ablated variants that lack objectness.  For instance, \textit{Ours} outperforms \textit{Ours-noObj} in 1-shot with gains of 1.4 and 2.2 percentage points in \textit{mIoU} and \textit{FB-IoU} respectively. Additionally, \textit{Ours-DENet} outperforms DENet~\cite{liu2020dynamic} by 1.5 and 3.1 percentage points in \textit{mIoU} for 1-shot and 5-shot segmentation respectively. 

For this dataset, we observe state-of-art performance with respect to \textit{mIoU} in both 1-shot and 5-shot results from our DENet variant that uses objectness: \textit{Ours-DENet}. In 1-shot experiments, \textit{Ours-DENet} outperforms all methods by at least 1.5 percentage points in \textit{mIoU}. In 5-shot, \textit{Ours-DENet} outperforms all methods by at least 3.1 percentage points in \textit{mIoU}.  


\begin{table}[!t]
\begin{center}
\begin{tabular}{| l | c  c  c  c | c | c | c  c  c  c | c | c |}
\hline
\multirow{3}{*}{Method}&\multicolumn{6}{c|}{1-shot}&\multicolumn{6}{c|}{5-shot}\\
\cline{2-13}
&\multicolumn{5}{c|}{\textit{mIoU}}&{\textit{FB-IoU}}&\multicolumn{5}{c|}{\textit{mIoU}}&{\textit{FB-IoU}}\\\cline{2-13}
& fold1 & fold2 & fold3 & fold4 & mean & mean & fold1 & fold2 & fold3 & fold4 & mean & mean\\
\hline
PANet~\cite{wang2019panet} & - & - & - & - & 20.9 & 59.2 &- & - & - & - & 29.7 & 63.5\\
FWB~\cite{nguyen2019feature} & - & - & - & - & 21.2 & - & - & - & - & - & 23.7 & - \\
SST~\cite{zhu2020self} & - & - & -  & - & 22.2 & - & - & - & - & - & 31.3 & -\\
DAN~\cite{wang2020few} & - & - & - & - & 24.4 & 62.3 & - & - & - & - & 29.6 & 63.9\\
PAPNet~\cite{liu2020part} & 36.5 & 26.5 & 26.0 & 19.7 & 27.2 & - & \textbf{48.9} & 31.4 & 36.0 & 30.6 & 36.7 & -\\
PMM~\cite{yang2020prototype} & 29.5 & 36.8 & 28.9 & 27.0 & 30.6 & - & 33.8 & 42.0 & 33.0 & 33.3 & 35.5 & -\\
PFENet~\cite{tian2020prior} & 33.4 & 36.0 & 34.1 & 32.8 & 34.1 & 60.0 & 35.9 & 40.7 & 38.1 & 36.1 & 37.7 & 61.6\\
DENet~\cite{liu2020dynamic} & 42.9 & 45.8 & \textbf{42.2} & 40.2 & 42.8 & - & 45.4 & 44.9 & 41.6 & 40.3 & 43.0 & -\\
\hline
Ours & 35.0 & 42.7 & 38.1 & 37.9 & 38.4 & 66.9 & 37.0 & 46.4 & 41.1 & 41.3 & 41.5 & \textbf{68.7}\\
Ours-noObj & 34.1 & 42.3 & 36.8 & 35.0 & 37.0 & 64.7 & 35.4 & \textbf{47.8} & 40.6 & 37.9 & 40.4 & 66.4\\
Ours-DENet & \textbf{48.0} & \textbf{46.9} & 41.6 & \textbf{40.8} & \textbf{44.3} & \textbf{67.7} & 48.8 & 47.5 & \textbf{43.7} & \textbf{44.5} & \textbf{46.1} & 68.3\\
\cline{1-13}
Ours-gtObj & 44.8 & 56.6 & 53.9 & 50.2 & 51.4 & 74.4 & 48.4 & 63.6 & 57.5 & 55.0 & 56.1 & 77.4\\
\hline
\end{tabular}
\end{center}
\vspace{0 pt}
\caption{Performance of the eight baselines and our approach on the task of 1-shot and 5-shot semantic segmentation for COCO-$20^i$. Ours-gtObj shows the upper bound by using ground truth objectness. (“-” means the original paper does not report performance for this metric.)}
\label{table:coco_1shot_supp}
\end{table}

\section{Qualitative Results in PASCAL-$5^i$} 
\begin{figure}[!th]
\centering
\includegraphics[width=0.98\textwidth]{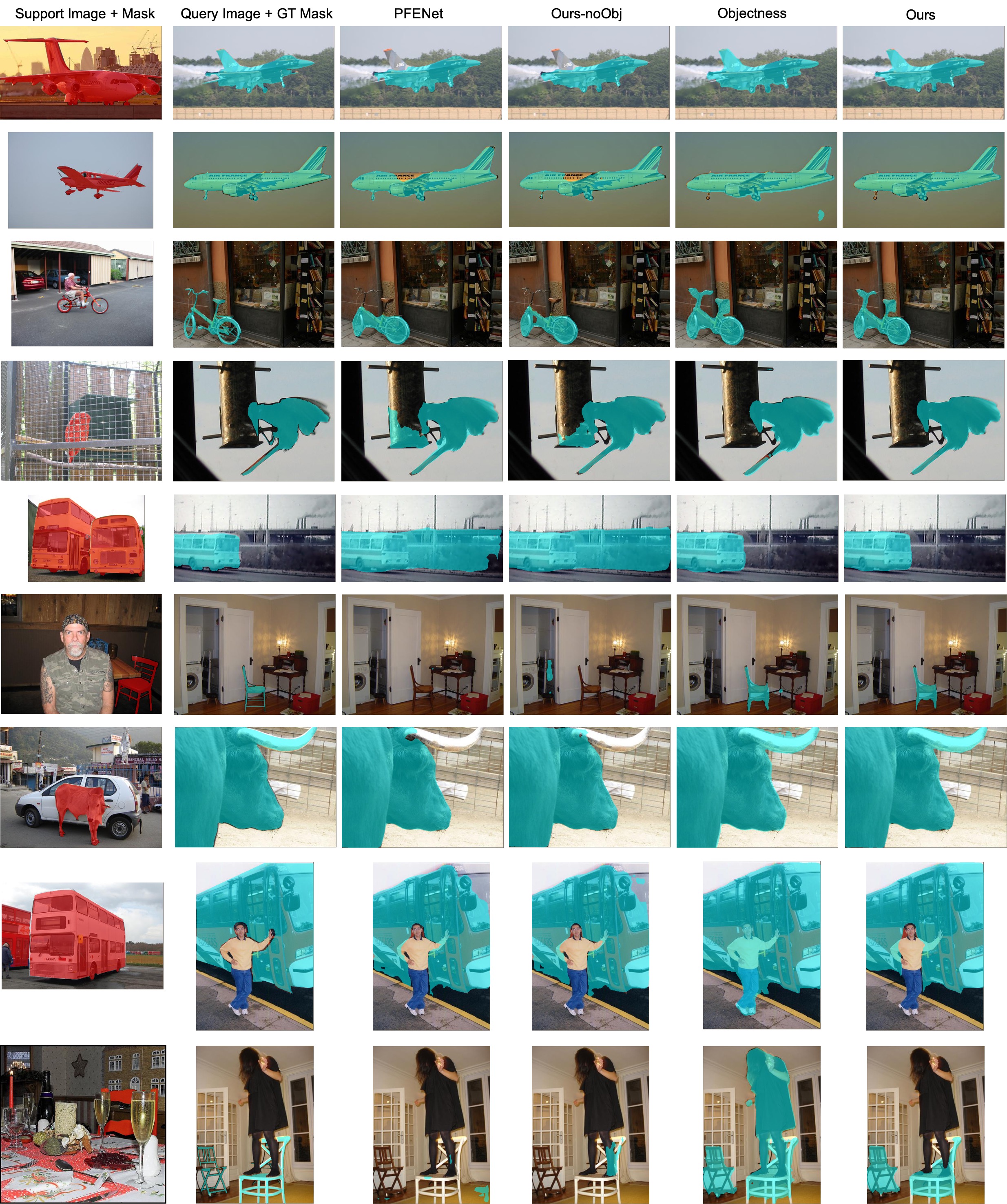}
\caption{Qualitative results for 1-shot semantic segmentation on PASCAL-$5^i$. From left to right we show the support image with the mask of the target category, the query image with ground truth segmentation for the target category, the prediction of PFENet~\cite{tian2020prior}, the prediction of our implemented baseline without objectness, the prediction of our objectness module and the prediction of our method with objectness. Qualitatively our method (column 6) produces more accurate segmentation than PFENet~\cite{tian2020prior} (column 3) and our implementation without objectness (column 4) by levaraging objectness (column 5), especially in the case of cluttered backgrounds and appearance mismatches between the support and query images.}
\label{fig:qualitative_pascal1}
\end{figure}

\begin{figure}[!th]
\centering
\includegraphics[width=0.95\textwidth]{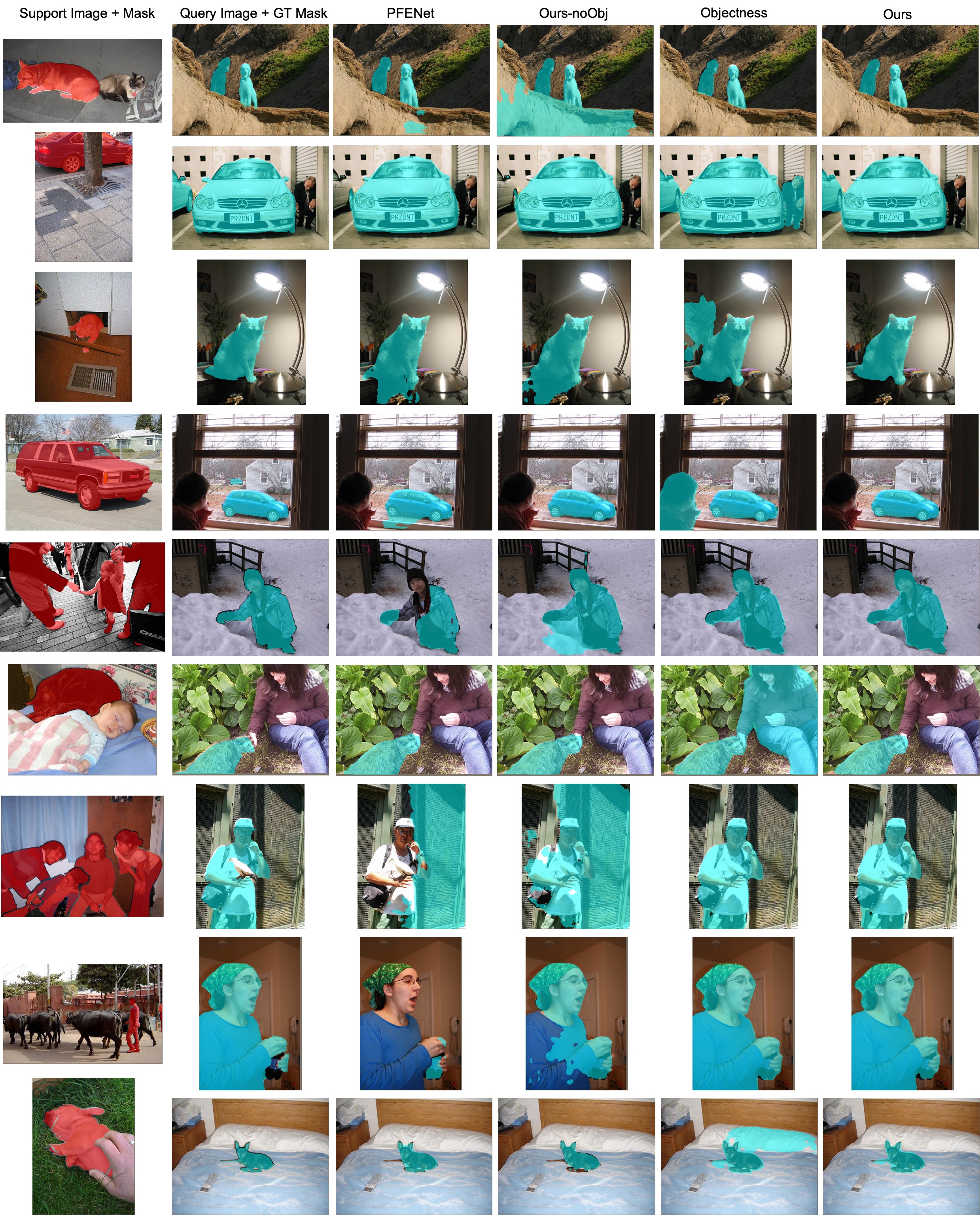}
\caption{Qualitative results for 1-shot semantic segmentation on PASCAL-$5^i$. From left to right we show the support image with the mask of the target category, the query image with ground truth segmentation for the target category, the prediction of PFENet~\cite{tian2020prior}, the prediction of our implemented baseline without objectness, the prediction of our objectness module and the prediction of our method with objectness. Qualitatively our method (column 6) produces more accurate segmentation than PFENet~\cite{tian2020prior} (column 3) and our implementation without objectness (column 4) by levaraging objectness (column 5), especially in the case of cluttered backgrounds and appearance mismatches between the support and query images.}
\label{fig:qualitative_pascal2}
\end{figure}

\begin{figure}[!th]
\centering
\includegraphics[width=0.98\textwidth]{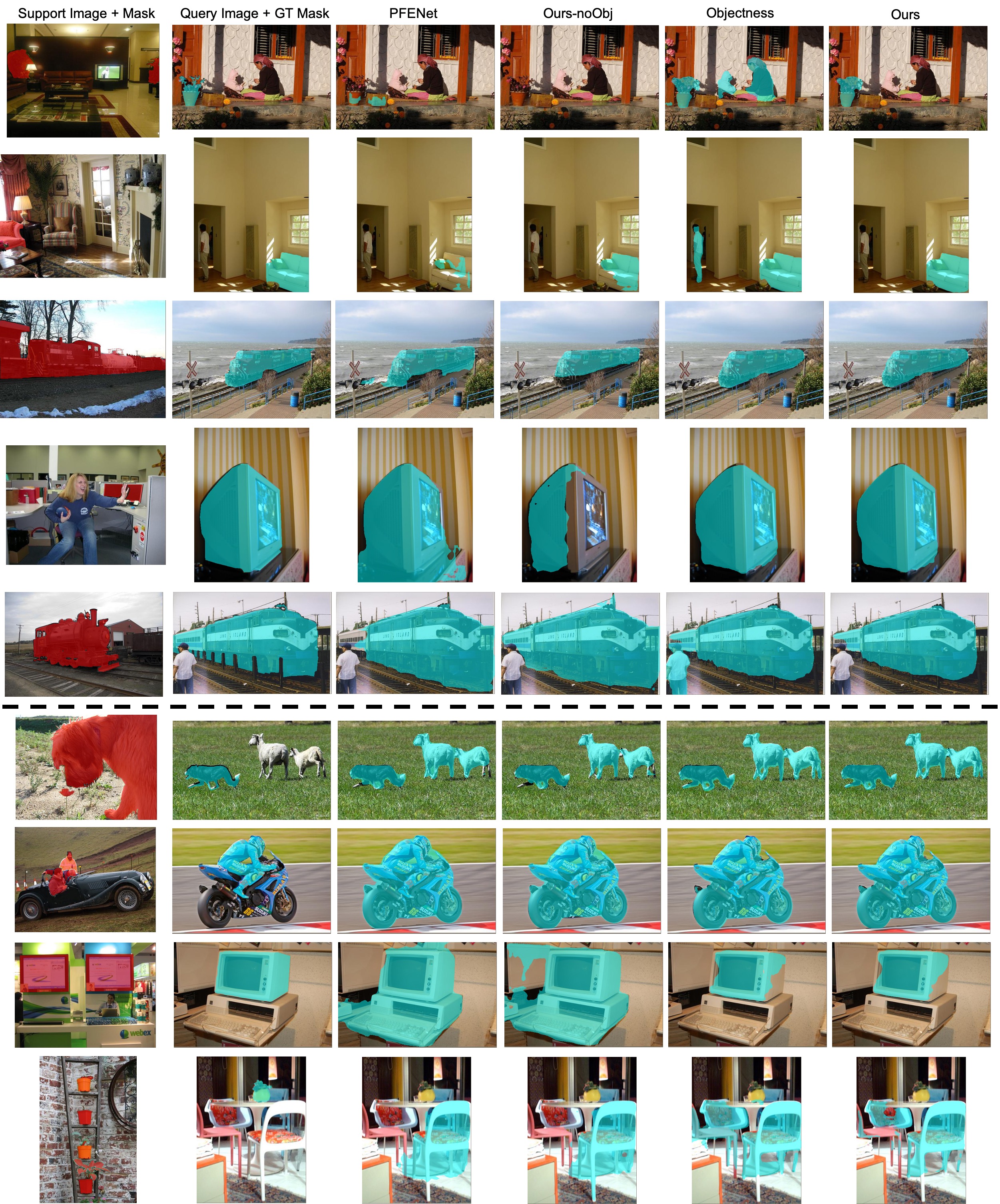}
\caption{Qualitative results for 1-shot semantic segmentation on PASCAL-$5^i$. From left to right we show the support image with the mask of the target category, the query image with ground truth segmentation for the target category, the prediction of PFENet~\cite{tian2020prior}, the prediction of our implemented baseline without objectness, the prediction of our objectness module and the prediction of our method with objectness. We show five success cases of our method in row 1-5 and four failure cases in row 6-9 (below the dotted line). In row 6-7, we show two failure cases of our objectness-aware comparison module. In rows 8-9, we show two failure cases of our objectness module.}
\label{fig:qualitative_pascal3}
\end{figure}

In Figures~\ref{fig:qualitative_pascal1}, \ref{fig:qualitative_pascal2}, and \ref{fig:qualitative_pascal3}, we show more qualitative 1-shot semantic segmentation results on PASCAL-$5^i$. Qualitatively our method (column 6) produces more accurate segmentations than PFENet~\cite{tian2020prior} (column 3) and our implementation without objectness (column 4). When the target object category has a large appearance mismatch between the support image and query image, PFENet~\cite{tian2020prior} (column 3) and \textit{Ours-noObj} (column 4) are usually limited to segmenting only part of the target region correctly, while our method (column 6) produces a more accurate segmentation by leveraging objectness (column 5) (row 1 of Figure~\ref{fig:qualitative_pascal1} as an example). When some background regions share similar features with the target object, PFENet~\cite{tian2020prior} (column 3) and \textit{Ours-noObj} (column 4) sometimes misclassify those background regions as the target object category, while our method (column 6) can correctly label them as negative by taking advantage of objectness (column 5) (row 1 of Figure~\ref{fig:qualitative_pascal2} as an example).

We also show the object regions predicted by our objectness module (column 5) to illustrate how our method benefits from leveraging objectness. When there is a single salient object in the query image (the aeroplane in row 1 of Figure~\ref{fig:qualitative_pascal1}, the bus in row 5 of Figure~\ref{fig:qualitative_pascal1}, etc.), our objectness module can predict an accurate object region for that object, and then our objectness-aware comparison module transfers the class-agnostic object prior to the final prediction. In these cases, the final prediction of our method looks almost the same as the prediction of our objectness module. When there are multiple objects in the image (row 4 of Figure~\ref{fig:qualitative_pascal2}, row 1 of Figure~\ref{fig:qualitative_pascal3}, etc.), our objectness module can highlight all the object regions while our objectness-aware comparison module removes irrelevant objects after observing class-specific features of the annotated object in the support image. Note that the object regions predicted by our objectness module can sometimes be noisy, but our objectness-aware comparison module has the ability to recover from such noise (row 2 and 7 in Figure~\ref{fig:qualitative_pascal1}). 

We show some failure cases in the bottom four rows of Figure~\ref{fig:qualitative_pascal3} (below the dotted line). In rows 8-9, we show two failure cases of our objectness module, i.e. in row 8 our objectness module produces an inaccurate segmentation of the monitor, and in row 9 it fails to recognize the plant. Our objectness-aware comparison module cannot recover from such errors of the objectness module in these two cases. We also observe a few cases in which our objectness module generates high-quality object regions but our comparison module 
mistakenly discards part of the target object after observing the class-specific features of the support object (the aeroplane in row 6 of Figure~\ref{fig:qualitative_coco_supp_1}). In row 6-7, we show additional failure cases where our objectness-aware comparison module fails to remove irrelevant objects, i.e. the sheep in row 6 and the motorcycle in row 7. This occurs probably because the irrelevant object shares similar features (produced by the feature extraction module) with the target object. By taking advantage of objectness, our approach typically can remove background regions which share similar features with the target object. However, it is still hard for our approach to remove irrelevant foreground object regions with similar features. A valuable direction for future work is to learn more discriminative features to differentiate such irrelevant object regions from the target object.

\section{Qualitative Results in COCO-$20^i$} 

\begin{figure}[!th]
\centering
\includegraphics[width=0.98\textwidth]{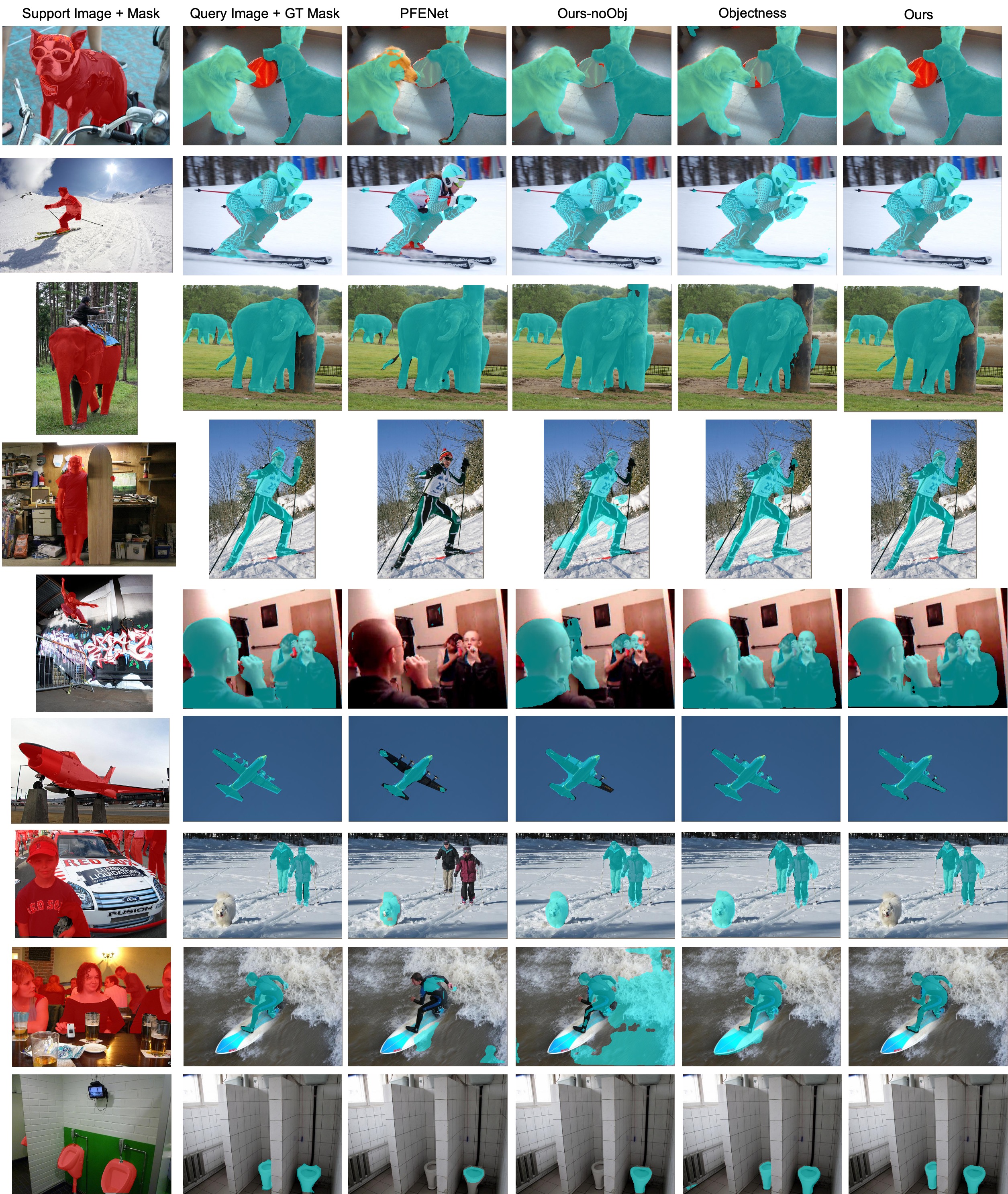}
\caption{Qualitative results for 1-shot semantic segmentation on COCO-$20^i$. From left to right we show the support image with the mask of the target category, the query image with ground truth segmentation for the target category, the prediction of PFENet~\cite{tian2020prior}, the prediction of our implemented baseline without objectness, the prediction of our objectness module and the prediction of our method with objectness. Qualitatively our method (column 6) produces more accurate segmentation than PFENet~\cite{tian2020prior} (column 3) and our implementation without objectness (column 4) by levaraging objectness (column 5), especially in the case of cluttered backgrounds and appearance mismatches between the support and query images.}
\label{fig:qualitative_coco_supp_1}
\end{figure}

\begin{figure}[!th]
\centering
\includegraphics[width=1.0\textwidth]{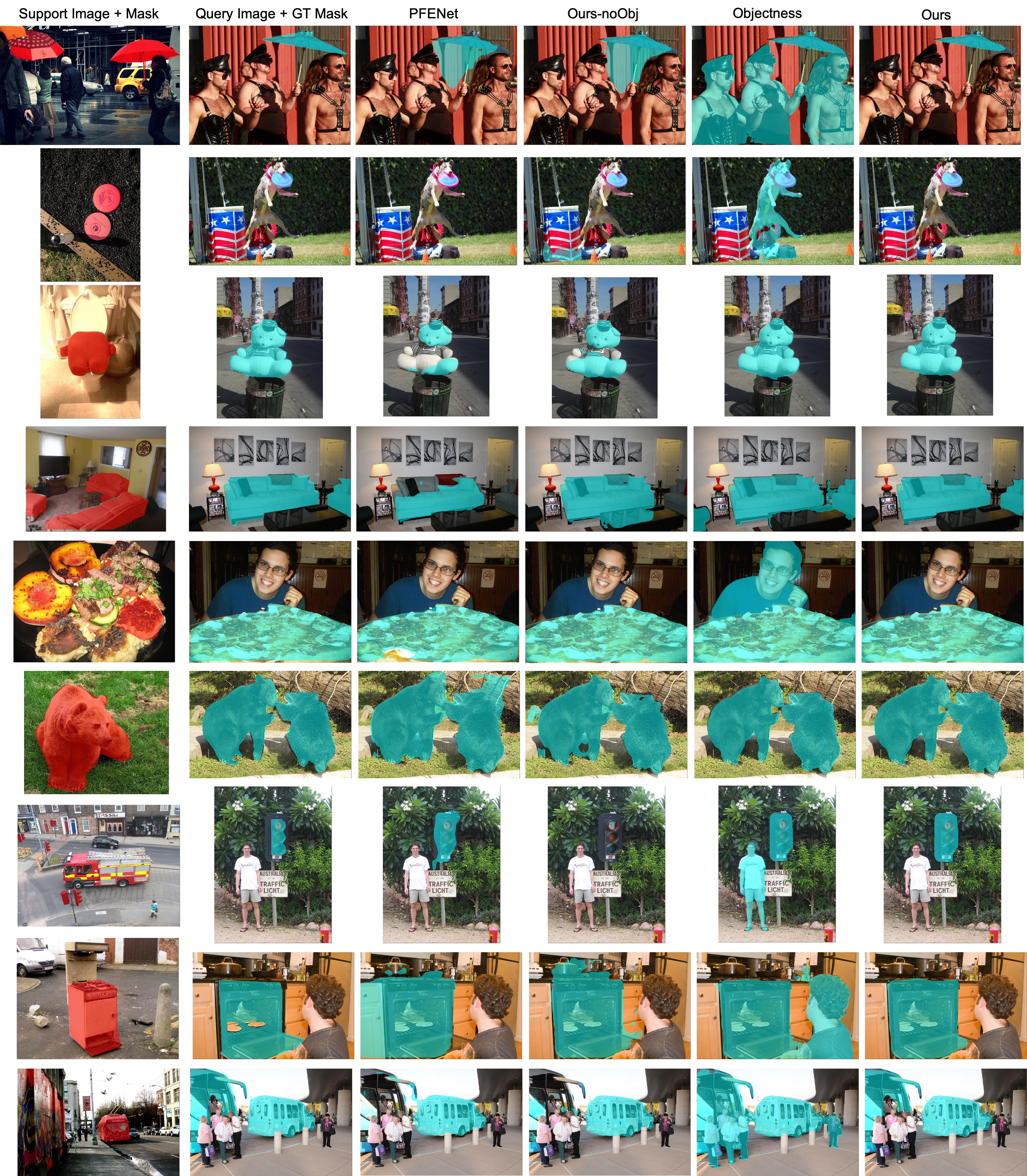}
\caption{Qualitative results for 1-shot semantic segmentation on COCO-$20^i$. From left to right we show the support image with the mask of the target category, the query image with ground truth segmentation for the target category, the prediction of PFENet~\cite{tian2020prior}, the prediction of our implemented baseline without objectness, the prediction of our objectness module and the prediction of our method with objectness. Qualitatively our method (column 6) produces more accurate segmentation than PFENet~\cite{tian2020prior} (column 3) and our implementation without objectness (column 4) by levaraging objectness (column 5), especially in the case of cluttered backgrounds and appearance mismatches between the support and query images.}
\label{fig:qualitative_coco_supp_2}
\end{figure}

\begin{figure}[!th]
\centering
\includegraphics[width=0.96\textwidth]{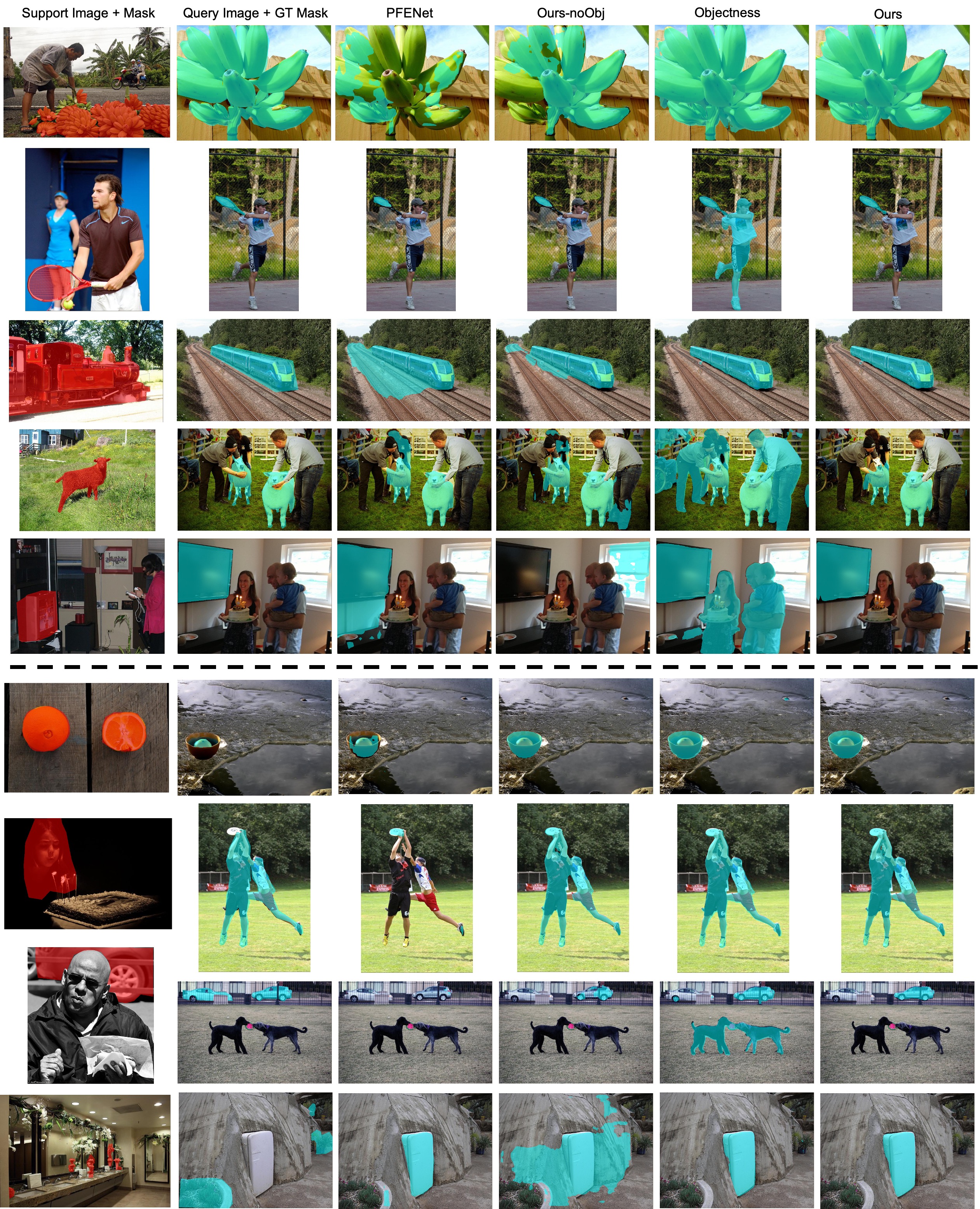}
\caption{Qualitative results for 1-shot semantic segmentation on COCO-$20^i$. From left to right we show the support image with the mask of the target category, the query image with ground truth segmentation for the target category, the prediction of PFENet~\cite{tian2020prior}, the prediction of our implemented baseline without objectness, the prediction of our objectness module and the prediction of our method with objectness. We show five success cases of our method in rows 1-5 and four failure cases in row 6-9 (below the dotted line). In rows 6-7, we show two failure cases of our objectness-aware comparison module. In rows 8-9, we show two failure cases of our objectness module.}
\label{fig:qualitative_coco_supp_3}
\end{figure}

We show more qualitative 1-shot semantic segmentation results on COCO-$20^i$ in Figures~\ref{fig:qualitative_coco_supp_1}, \ref{fig:qualitative_coco_supp_2}, and \ref{fig:qualitative_coco_supp_3}. Overall, our method (column 6) produces more accurate segmentations than PFENet~\cite{tian2020prior} (column 3) and our implementation without objectness (column 4), especially in the case of cluttered backgrounds and appearance mismatches between the support and query images. As in PASCAL-$5^i$, we also show failure cases in the bottom 4 rows of Figure~\ref{fig:qualitative_coco_supp_3} (below the dotted line), where rows 8-9 show two failure cases of the objectness module and rows 6-7 show two failure cases of the objectness-aware comparison module. Our method enables a separation of class-agnostic and class-specific features. It is valuable to explore in future work how to learn more effective class-agnostic features to address the errors of our objectness module, and more generalizable class-specific features to address the failure cases of our objectness-aware comparison module.

\section{Parallel Fine-Grained Analysis}
In the main paper, we report our fine-grained analysis with respect to \textit{Feature extraction module} on PASCAL-$5^i$ in Section 4.3. In parallel, we also conduct similar analysis of the \textit{Feature extraction module} on COCO-$20^i$ to reinforce our findings, as shown in Table~\ref{table:coco_1shot_feat}. In rows 1 and 2,  we report results for two implementations without objectness where the feature extraction module relies on  ResNet-101~\cite{he2016deep} and HRNetV2-W48~\cite{sun2019high}. In rows 3 and 4, we report results for two  implementations of our method that use HRNetV2-48 as the backbone of the objectness module. 

With objectness (row 3,4), the better \textit{mIoU} and \textit{FB-IoU} of HRNetV2-W48 compared to ResNet-101 reinforces our findings in PASCAL-$5^i$ that HRNetV2-W48 projects images into a better feature space than ResNet-101 when leveraging objectness. The consistent gains with the objectness module reinforce the advantage of introducing it in few-shot segmentation; e.g. 1.3 percentage point gain in \textit{mIoU} and 1.0 in \textit{FB-IoU} for ResNet-101 (row 1,3) and 3.9 percentage point gain in \textit{mIoU} and 3.6 in \textit{FB-IoU} for HRNetV2-48 (row 2,4). 

\setlength{\tabcolsep}{4.5pt}
\begin{table}[!t]
\begin{center}
\begin{tabular}{| l | c c c c | c | c |}
\hline
\multirow{2}{*}{Feature}&\multicolumn{5}{c|}{\textit{mIoU}}&{\textit{FB-IoU}}\\\cline{2-7}
& fold1 & fold2 & fold3 & fold4 & mean & mean\\
\hline
Res101* & 28.4 & 22.4 & 18.9 & 15.6 & 21.3 & 59.8\\
HRNet* & 22.5 & \textbf{25.1} & 19.1 & 8.4 & 18.8 & 57.8\\
\hline
Res101 & 29.6 & 22.9 & \textbf{20.3} & \textbf{17.5} & 22.6 & 60.8\\
HRNet & \textbf{31.6} & 24.1 & 19.4 & 15.6 & \textbf{22.7} &  \textbf{61.4}\\
\hline
\end{tabular}
\end{center}
\vspace{-6pt}
\caption{\textit{mIoU} and \textit{FB-IoU} with multiple backbones as the feature extraction module for 1-shot segmentation on COCO-$20^i$. Results without objectness are in rows 1-2. In rows 3-4, we show results with HRNetV2-48 as the backbone of the objectness module.}
\vspace{-1pt}
\label{table:coco_1shot_feat}
\end{table}

\end{document}